\documentclass[letterpaper, 10 pt, conference]{ieeeconf}  

\IEEEoverridecommandlockouts                              

\usepackage{graphics} 
\usepackage{subfigure}
\usepackage{amsmath}
\usepackage{amssymb}
\usepackage{array}
\usepackage{multirow}
\usepackage[pdftex]{graphicx}
\usepackage{eqparbox}
\usepackage{float}
\usepackage{booktabs}
\usepackage{tabu}
\usepackage{dsfont}
\usepackage{color, colortbl}
\usepackage{booktabs, cellspace, multirow, tabularx}
    \newcolumntype{R}{>{\raggedright\arraybackslash}X}
    \addparagraphcolumntypes{R}
\usepackage{ragged2e}

\usepackage[table]{xcolor}
\usepackage[export]{adjustbox}
\usepackage{amsmath}

\usepackage[colorlinks,allcolors=blue]{hyperref}

\title{\LARGE \bf
Semantic-Supervised Spatial-Temporal Fusion for LiDAR-based 3D Object Detection
}

\author{
Chaoqun Wang$^{1,2*}$\thanks{*Equal contribution.}, 
\thanks{$^{1}$ Sun Yat-sen University.}
\thanks{$^{2}$ The Chinese University of Hong Kong, Shenzhen. chaoqunwang@link.cuhk.edu.cn}
Xiaobin Hong$^{3*}$, 
\thanks{$^{3}$ Nanjing University. \{xiaobinhong@smail.lwz@\}nju.edu.cn}
Wenzhong Li$^{3}$, 
and
Ruimao Zhang$^{1\dag}$\thanks{\dag Corresponding author. ruimao.zhang@ieee.org}\\
}
\begin{document}

\maketitle
\thispagestyle{empty}
\pagestyle{empty}

\begin{abstract}
LiDAR-based 3D object detection presents significant challenges due to the inherent sparsity of LiDAR points. A common solution involves long-term temporal LiDAR data to densify the inputs. However, efficiently leveraging spatial-temporal information remains an open problem.
In this paper, we propose a novel Semantic-Supervised Spatial-Temporal Fusion~(ST-Fusion) method, which introduces a novel fusion module to relieve the spatial misalignment caused by the object motion over time and a feature-level semantic supervision to sufficiently unlock the capacity of the proposed fusion module.
Specifically, the ST-Fusion consists of a Spatial Aggregation (SA) module and a Temporal Merging~(TM) module. The SA module employs a convolutional layer with progressively expanding receptive fields to aggregate the object features from the local regions to alleviate the spatial misalignment, the TM module dynamically extracts object features from the preceding frames based on the attention mechanism for a comprehensive sequential presentation. Besides, in the semantic supervision, we propose a Semantic Injection method to enrich the sparse LiDAR data via injecting the point-wise semantic labels, using it for training a teacher model and providing a reconstruction target at the feature level supervised by the proposed object-aware loss.
Extensive experiments on various LiDAR-based detectors demonstrate the effectiveness and universality of our proposal, yielding an improvement of approximately +2.8\% in NDS based on the nuScenes benchmark.
\end{abstract}

\section{INTRODUCTION}
LiDAR-based 3D object detection is pivotal in many real-world applications, including autonomous driving and robotics~\cite{zhou2018voxelnet,yan2018second,lang2019pointpillars,yin2021center}. Traditionally, most existing methods treat individual LiDAR frames independently, such manners neglect the power of sequential LiDAR data, which contain valuable spatiotemporal information that can greatly densify the sparse points and enhance detection performance.

\begin{figure*}[tp]
    \centering
    \includegraphics[width=1.0\linewidth]{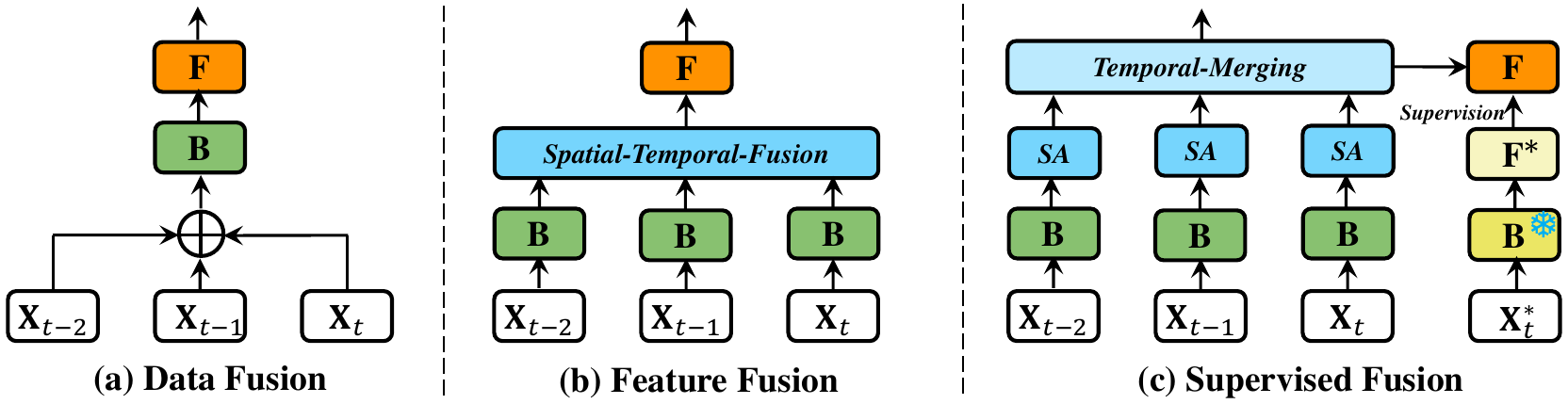}
    \caption{Sequential LiDAR fusion methods. (a). Data-level fusion. The sequential LiDAR data is directly stacked. (b). Feature-level fusion. A designed spatial-temporal fusion model fuses sequential LiDAR features. (c). Ours semantic-supervised spatial-temporal fusion~(ST-Fusion), consists of a Spatial Aggragate~(SA) module and a Temporal Merging~(TM) module. The enriched semantically LiDAR data guides the spatial-temporal fusion module learning via feature-level supervision. Here $\textbf{B}$, $\textbf{F}$ are the backbone and features respectively. $\textbf{X}^{*}$ is the enriched semantically LiDAR data.}
    \label{fig:intro}
    \vspace{-3mm}
\end{figure*}

To this end, numerous approaches sprung up to leverage sequential LiDAR for robust 3D object detection. The early works~\cite{lang2019pointpillars,yin2021center,qi2021offboard} attempted to exploit temporal information by directly stacking LiDAR points as shown in Fig.~\ref{fig:intro}-(a), but such rough solutions often struggle with long-term sequential LiDAR data with complex scenes and object motions. In contrast, as shown in Fig.~\ref{fig:intro}-(b), some works~\cite{lin2022sparse4d,liu2023stformer3d,fang2023tsc} elaborately designed spatial-temporal fusion modules to handle the sequential data and obtain feature representation by harnessing the learning ability of the deep models. However, such approaches always neglect well-crafted supervision for the fusion modules, which limits its capacity.

To address the aforementioned issues, we propose a novel sequential-LiDAR-based 3D object detection architecture as shown in Fig.~\ref{fig:intro}-(c), termed Semantic-Supervised Spatial-Temporal Fusion~(ST-Fusion). ST-Fusion consists of a Spatial Aggregation~(SA) and a Temporal Merging~(TM) module. The SA mainly mitigates the spatial misalignment by aggregating the spatial features from local regions, while the TM module dynamic queries the features from the preceding frame for a more comprehensive representation. Additionally, we proposed a Semantic-Supervised Training~(SST) method, which provides feature-level supervision to sufficiently unlock the potential of the fusion module. 

Specifically, the SA module addresses feature shifts caused by object motion over time by using a simple convolutional layer, with a progressively expanding kernel size to adaptively capture the features from neighboring regions.
In the TM module, pixel-wise features are dynamically extracted from preceding frames through an attention mechanism.
For the SST, we introduce a Semantic Injection method to enhance the sparse LiDAR data. Such enhanced semantically data involves training a teacher model to generate robust features, which are the reconstruction targets of our proposed object-aware supervision. In practice, Semantic Injection classifies point labels based on ground truth and incorporates them as an additional dimension in the LiDAR points.

To demonstrate the its effectiveness, we conducted experiments based on variant detectors, including PointPillar~\cite{lang2019pointpillars}, SECOND~\cite{yan2018second}, CenterPoint~\cite{yin2021center}, and TransFusion-L~\cite{bai2022transfusion}, and universally gains huge promotions on the nuScenes. In practice, our proposal achieved +3.30\%, +3.18\%, +2.74\%, and +1.93
\% increments in NDS respectively. These promotions highlight the effectiveness and universality of our proposal in boosting LiDAR-based 3D object detection accuracy.

The main contributions are summarized as follows.
\begin{itemize}
     \item We propose a novel Spatial-Temporal Fusion module, which introduces a progressive Spatial Aggregation~(SA) module to relieve object motion over time and an attention-based Temporal Merging~(TM) module to dynamically fuse sequential features.

    \item We propose a Semantic-Supervised training pipeline, consisting of a Semantic Injection method to enrich the semantic information into LiDAR points, and an object-aware supervision to optimize the fusion module.
    
    \item We make extensive experiments on multiple famous detectors, and gain around $2.8\%$ NDS improvements on the nuScence benchmark to demonstrate the effectiveness and universality of our proposal. 
\end{itemize}

\section{Related Work}

\subsection{LiDAR-based Detection}
\label{sec:detection}

LiDAR-based 3D Object Detection focuses on localizing and classifying 3D objects within a point cloud scene. There are three primary approaches in this domain. The first approach processes raw point clouds directly, employing networks like PointNet++~\cite{qi2017pointnet++} to extract point-wise features and generate 3D proposals, as seen in~\cite{shi2019pointrcnn,chen2019fast,yang2019std,zhang2022not}. The second approach leverages a pillar encoding mechanism~\cite{lang2019pointpillars,wang2020pillar}, which converts the point cloud into a grid of pillars. These pillar-wise features are then processed using 2D convolutional networks to extract bird’s-eye view (BEV) features, which are subsequently fed into a detection head. The third and most widely adopted approach, particularly in autonomous driving scenarios, involves voxelization of the point cloud~\cite{zhou2018voxelnet,yan2018second,shi2020pv}. Voxel-wise features are extracted using 3D sparse convolutional networks \cite{3DSemanticSegmentationWithSubmanifoldSparseConvNet}, enabling efficient and accurate detection. More recently, an alternative perspective for detection has been proposed, departing from traditional field-of-view and bird’s-eye view approaches. These methods, such as those introduced by \cite{bewley2021range,sun2021rsn}, utilize range view representations to tackle the object detection task, offering new insights into handling 3D point cloud data.

\subsection{Spatial-Temporal Fusion}
\label{sec:st-related}

Spatial-temporal fusion is pivotal in overcoming the core challenges in LiDAR-based 3D object detection for autonomous driving, particularly those associated with sparse data, occlusion, and complex object motion. Earlier methods~\cite{qi2021offboard, yan2018second,piergiovanni20214d} sought to leverage temporal information by directly stacking consecutive LiDAR frames. While this approach provided a straightforward solution for incorporating temporal data, it often struggled to handle the complexities of long-term sequences, especially when dealing with dynamic environments and rapidly moving objects. 
To overcome this, recent works~\cite{xu2022int,koh2023mgtanet} have proposed learning-based scene flow estimation from dynamic point clouds through end-to-end deep neural networks. Additionally, other research efforts introduced sophisticated spatial-temporal fusion architectures to address specific challenges like data sparsity~\cite{lin2022sparse4d,hou2024query}, occlusion~\cite{liu2023stformer3d,chen2022mppnet}, and object motion~\cite{park2024lidar,hou2024query} by leveraging both spatial and temporal domains. However, these approaches often lack well-designed feature-level supervision, which can limit the potential of fusion models.

\subsection{Knowledge Distillation}

Knowledge distillation (KD) \cite{hinton2015distilling} has become a valuable technique for improving deep learning models, especially in resource-limited environments. In 3D object detection, KD enables the transfer of knowledge from a larger teacher model to a smaller student model, enhancing detection accuracy without significant computational overhead \cite{gou2021knowledge,zhu2021student}. Early methods adapted KD from 2D detection to 3D space \cite{yang2022towards,li2023representation,cho2023itkd}, distilling intermediate features or final outputs from 2D image-based models to 3D point cloud detectors, allowing students to learn richer representations.
As the field evolved, specialized KD techniques emerged to address challenges unique to 3D data. Methods began focusing on distilling backbone features \cite{romero2014fitnets}, attention mechanisms \cite{zagoruyko2016paying,zhang2020improve}, and relational structures \cite{li2020semantic,park2019relational,tung2019similarity}, helping students better capture task-specific knowledge. Multi-scale feature distillation has also been introduced to ensure critical contextual information is transferred across various scales and viewpoints \cite{wang2020multi,jiang2024fsd}. More recently, hybrid approaches have integrated KD with self-supervised and contrastive learning to improve the model’s generalization, especially in scenarios with limited labeled 3D data \cite{zhang2023qd}. These hybrid methods strengthen feature learning and reduce reliance on labeled datasets, making them highly applicable in real-world settings where annotation is costly. 
Despite these advancements, optimizing the KD process for the inherently high-dimensional and complex nature of 3D data remains challenging and needs continuous efforts.

\begin{figure*}
    \centering
    \includegraphics[width=0.99\linewidth]{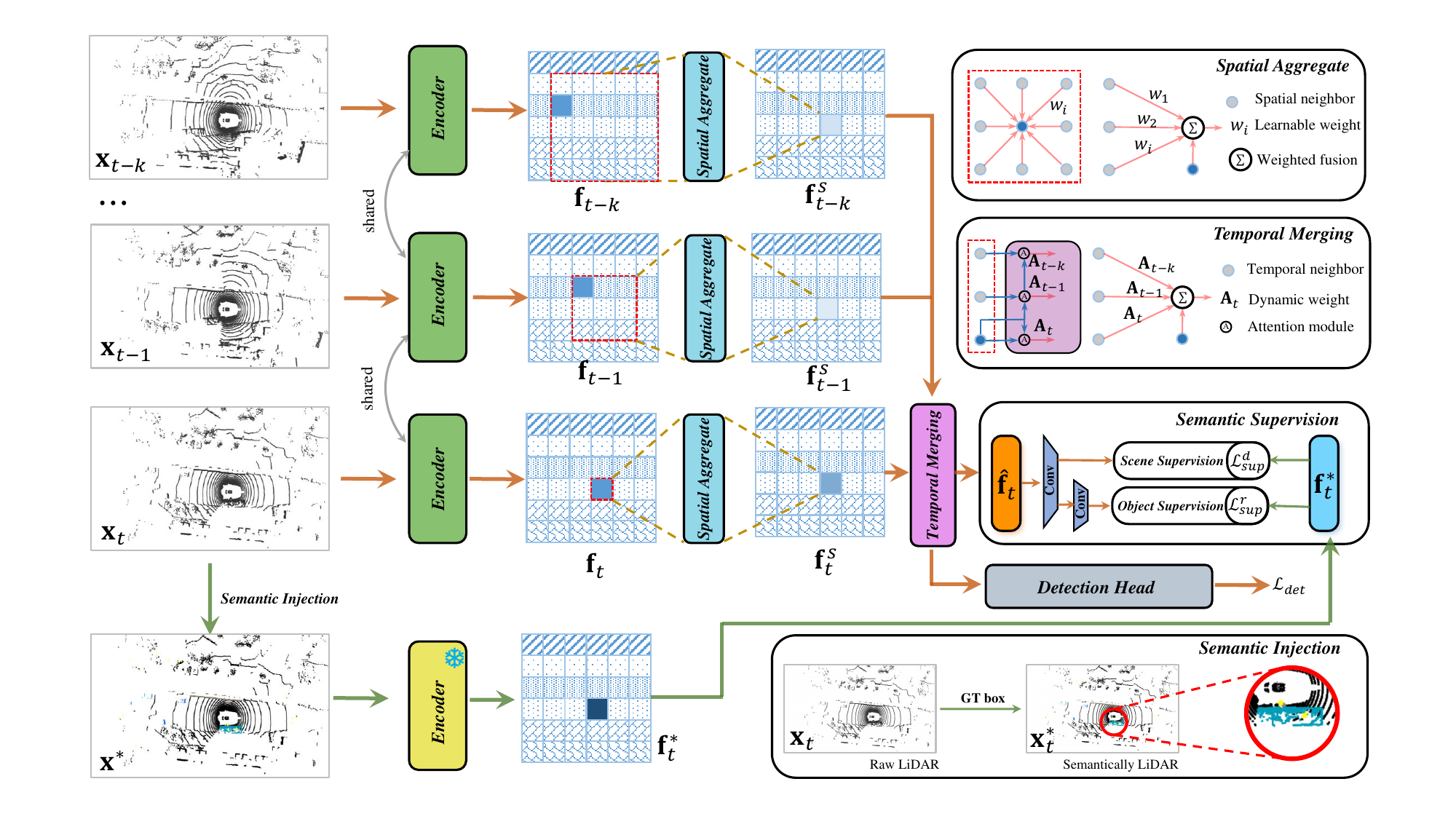}
    \vspace{-3mm}
    \caption{The overall framework of our proposed Semantic-Supervised Spatial-Temporal Fusion. For a given sequential LiDAR data, we feed it into a shared encoder and obtain the fusion feature $\hat{\textbf{f}}_t$ via the proposed Spatial and Temporal fusion module, which is fed into the detection head for the final prediction. Besides, we inject the semantic class labels of each point into the sparse LiDAR data via the proposed Semantic Injection module and gain high-quality features $\textbf{f}^{*}_t$ with a freeze backbone used for object-aware Semantic Supervision. The whole model is supervised by a simple detection loss $\mathcal{L}_{det}$ and proposed semantic supervision $\mathcal{L}_{sup}$, which consists of a scene distillation loss $\mathcal{L}^{d}_sup$ and object-aware reconstruction loss $\mathcal{L}^{r}_{sup}$.}
    \label{fig:pipeline}
\vspace{-3mm}
\end{figure*}

\section{Method}
\label{sec:method}

\subsection{Overall Framework}
\label{sec:framework}
Given a detector $\mathbf{D}$, consisting of an encoder and a detection head. 
Let $(\mathbf{x}_t, \mathbf{y}_t) \in \mathcal{D}$ denote the raw point cloud data in $t$ timestamp and its corresponding ground truth. The input $\mathbf{x}_t$ is fed into the encoder to extract the feature representation $\mathbf{f}_t$, which is then passed through the detection head to produce the final predictions.
In this paper, we propose a Semantic-Supervised Spatial-Temporal Fusion method to expand the single-frame input of a detector into multi-frame sequential inputs as shown in Fig.~\ref{fig:pipeline}.

Specifically, for $k$ frames input $\{\mathbf{x}_{t-k}, \mathbf{x}_{t-k+1}, \cdots, \mathbf{x}_{t} \}$, we feed the sequential inputs into the shared LiDAR encoder and obtain corresponding deep features $\{\mathbf{f}_{t-k}, \mathbf{f}_{t-k+1}, \cdots, \mathbf{f}_{t}\}$. To alleviate the feature misalignment in spatial space caused by the object motion over time, we propose a Spatial Aggregation~(SA) module to aggregate the local features from progressively expanding receptive fields $\{\mathbf{f}^{s}_{t-k}, \mathbf{f}^{s}_{t-k+1}, \cdots, \mathbf{f}^{s}_{t}\}$. Meanwhile, to leverage the temporal data, we propose a novel attention-based Temporal Merging~(TM) module that extracts features from the multiple preceding frames and merges them together as the fusion feature $\hat{\textbf{f}}_t$.

To fully leverage the proposed Spatial-Temporal Fusion module, we propose a Semantic-Supervised Training (SST) pipeline.
In practice, we introduce a novel Semantic Injection module, which injects the semantic class labels of each point into the sparse LiDAR data, $\mathbf{x}^{*}_t \leftarrow \mathcal{S}(\mathbf{x}_t, \mathbf{y}_t)$. Then, the enhanced semantically LiDAR data is used to train a strong teacher model and produce highly representational features $\mathbf{f}^{*}_t$, as the reconstruction target of the fused features at the feature level. The proposed object-aware semantic supervision is used to optimize the fusion module learning with a simple knowledge distillation strategy.

In summary, the training objective function could be formulated as follows:
\begin{equation}
\begin{aligned}
    \mathcal{L} = \mathcal{L}_{det}(\textbf{D}({\mathbf{x}_{t-k}, \cdots, \mathbf{x}_{t}}, \theta), \mathbf{y}_t) + \lambda \mathcal{L}_{sup}(\hat{\mathbf{f}}_t, \mathbf{f}^{*}_t 
    ) \\
\end{aligned}
\end{equation}

where $\textbf{D}$ is the detector and $\theta$ is the corresponding learnable parameters. Here the $\lambda$ is the tread-off weight and we set $\lambda=0.1$ in our experiments. $\mathcal{L}_{det}, \mathcal{L}_{sup}$ are the detection loss and proposed semantic supervision loss respectively.

\subsection{Spatial-Temporal Fusion}
\label{sec:st-fusion}

In this section, we introduce the proposed Spatial-Temporal Fusion method, which consists of a Spatial Aggregate~(SA) module to relieve the spatial misalignment and a Temporal Merging~(TM) module to extract comprehensive features from sequential inputs. 

Given a sequential LiDAR inputs $\{\mathbf{x}_{t-k}, \mathbf{x}_{t-k+1}, \cdots, \mathbf{x}_{t} \}$, we firstly feed it into a shared LiDAR encoder and obtain the corresponding features $\{\mathbf{f}_{t-k}, \mathbf{f}_{t-k+1}, \cdots, \mathbf{f}_{t}\}$, where $\textbf{f}_i \in \mathbb{R}^{C \times H \times W}$ is the 2D features as the $t$ timestamp. The sequential features inevitably misaligned for an object over time caused by its motion. To alleviate the above drawback, we propose a Spatial Aggregation module, which employs a convolutional layer to generate the object representation from neighbors. Intuitively, objects at earlier times have a longer range of motion, here we novelly utilize a progressively expanding kernel size to exploit a larger reception field. Here the SA module could be formulated as:
\begin{align}
\mathbf{f}_{t-i}^{s} = \mathbf{f}_{t-i} + \sigma(\mathcal{C}(\textbf{f}_{t-i}, \textbf{W}, m)),
\end{align}
where $\mathcal{C}$ is a convolutional layer followed by an activate function $\sigma$. Considering the objects' motion,
the $m=2i+1$ is the manually set progressively expanding kernel size, and $\textbf{W}$ is the learnable parameter.

Temporal Merging module terms to fuse multi-frame temporal features through an attention-based query function. 
Given a sequential spatial feature $\{\mathbf{f}^{s}_{t-k}, \mathbf{f}^{s}_{t-k+1}, \cdots, \mathbf{f}^{s}_{t}\}$, our object is to obtain the fused feature $\hat{\textbf{f}}_t$ that encapsulates temporal dependencies across $k$ previous frames while maintaining the spatial integrity of the $t$-th frame. To achieve it, we compute a temporal coefficient $\textbf{A} \in \mathbb{R}^{k \times H \times W}$, where $\textbf{A}_i \in \mathbb{R}^{H \times W}$ indicates the pixel-wise coefficient of features at $(t-i)$-th frame compared with the current frame. The normalized weight could be computed as:
\begin{align}
    \textbf{A}_{i} = \frac{\exp((\mathbf{f}^{s}_t\parallel \mathbf{f}^{s}_{t-i})\mathbf{W}_{a}^{\top})}{\sum_{j=1}^{k}\exp((\mathbf{f}^{s}_t\parallel \mathbf{f}^{s}_{t-j})\mathbf{W}_{a}^{\top})},
\end{align}
where $\parallel$ is the concatenate operation,  $\mathbf{W}_a\in \mathbb{R}^{H \times W\times 2C}$ denotes the learnable projection and $i=1, \cdots, k$. Therefore, the query function takes the correlation between each preceding and the current frame into account, and the fused representation is finally aggregated as follows:
\begin{align}
\label{eq:tm}
    \hat{\mathbf{f}}_{t} = \mathbf{f}^{s}_{t} + \sum_{i=1}^{k}\textbf{A}_i \odot \mathbf{f}^{s}_{t-i},
\end{align}
where $\odot$ is the element-wise Hadamard product.

\subsection{Semantic Supervision}
\label{sec:semantic}
As shown in Fig.~\ref{fig:pipeline}, semantic supervision mainly consists of two parts, including Semantic Injection and object-aware Semantic Supervision. Here we omit the timestamp $t$ for a clearer description.

\textbf{Semantic Injection.} Given sparse single-frame LiDAR data $\mathbf{x} = 
\{x_1, x_2, ..., x_n\}$, where $x_i \in \mathbb{R}^d$ indicates the $i$-th LiDAR point with $d$ dimension, Semantic Injection mainly injects class label of each point obtained from the ground-truth bounding box as well as the corresponding label, and obtains the semantically LiDAR data $\textbf{x}^{*}_t = \{x^{*}_1, x^{*}_2, ..., x^{*}_n\}$, where $x^{*}_i = (x_i \parallel c_i)$ and $c_i$ indicates the label of the $i$-th point. Here class label $c_i$ is computed as:
\begin{equation}
c_i=\left\{
\begin{aligned}
  i, ~\mathcal{G}(x_i)=i,\\
  0, ~\text{otherwise},
\end{aligned}
\right.
\end{equation}
where $\mathcal{G}(x_i)$ means the point $x_i$ belongs to a bounding box in the ground truth and $i$ is the corresponding class label, such as car, pedestrian, and so on. For the points of background, we set $c_i=0$.

\textbf{Semantic Supervision} mainly focuses on supervising the fusion module learning through an object-aware loss, which aims to mimic the high-quality semantic-riched features from the enhanced semantically LiDAR data and the sequential LiDAR data. Given the features from ST-Fusion module $\hat{\textbf{f}}$ and the corresponding semantic-riched features $\textbf{f}^{*}$,
the supervision mainly consists of two parts, including the scene and object reconstruction loss.

Scene distillation loss distills all of the regions in the scene equally. 
Following the previous methods~\cite{romero2014fitnets}, 
we transfer the feature representation of the baseline model to the same space with the high-quality features by $\mathbf{f}' \leftarrow \phi_1(\hat{\mathbf{f}})$, 
and then minimize the L2 distance within the features as follows,
\begin{align}
    \mathcal{L}_{sup}^d(\mathbf{f}', \mathbf{f}^{*}) &= \frac{1}{|\mathbb{N}|} \sum_{(i,j)\in \mathbb{N}} ||~\mathbf{f}'^{(i,j)}-\mathbf{f}^{*(i,j)}~||^2,
\end{align}
where $\mathbb{N}=\{(i,j) | 1\leq i \leq H, 1\leq j \leq W\}$ indicates the positions on the features,
and $|\mathbb{N}|=H \times W$ is the number of positions.
$\mathbf{f}'^{(i,j)}$ and $\mathbf{f}^{*(i,j)}$ denote the feature at the position $(i,j)$ respectively, and $\phi_1$ is a convolutional layer with the kernel size set to $1\times 1$.

Object reconstruction loss aims to mimic the feature representation at the object region.
Since the objects occupy a small part compared with the background, the above scene distillation can hardly supervise reconstructing the object representation from the sequential data.
To address the above issue, we come up with a sequential convolution decoder to distill the semantic prior from the high-quality semantic-rich features and minimize the feature reconstruction loss. 
As shown in Fig.~\ref{fig:pipeline}, we first transfer the student feature response to the semantic-rich feature space as $\mathbf{f}' \leftarrow \phi_1(\mathbf{f})$, 
sharing the parameters with the above scene distillation. 
Secondly, we further mine the neighborhood information via a sequential decoder: $\mathbf{f}^{\dag} \leftarrow \phi_2 (\mathbf{f}^{'}, m, n)$, 
where $\phi_2(\cdot, m, n)$ indicates a sequential decoder block with $m$ layers, and $n \times n$ kernels for each.
Finally, we minimize the object-relevant feature reconstruction loss as follows,

\begin{align}
    \mathcal{L}_{sup}^r(\mathbf{f}^{\dag}, \mathbf{f}^{*}) &= \frac{1}{|\mathbb{N}|}\sum_{(i,j)\in\mathbb{N}} ||~\mathbf{f}^{\dag (i,j)} -\mathbf{f}^{*(i,j)}~||^2 *\mathcal{W}(i,j,\mathbf{y}), \\
    \mathcal{W}(i,j,\mathbf{y}) &= \max \limits_{b\in \mathbf{y}}~ \text{exp}(\frac{(b_x-i)^2 + (b_y-j)^2}{2\sigma^2}),
\end{align}

\begin{table*}[t]
\caption{The experiment results~(mAP~$\%$, NDS, and the final mAP~$\%$ for each class) on nuScenes validation set. 'C.V.', 'Ped.', and 'T.C.' are short for construction vehicle, pedestrian, and traffic cone.}
\vspace{-3mm}
\centering
\setlength{\tabcolsep}{2mm}{
\renewcommand\arraystretch{1.3}
\resizebox{2.0\columnwidth}{!}{
\begin{tabular}{ccccccccccccc}
\toprule[1pt]
Detector & Car & Truck & C.V. & Bus & Trailer & Barrier & Motor. & Bike & Ped. & T.C. & mAP & NDS \\

\arrayrulecolor{black}\hline

\multirow{2}{*}{PointPillars}  & 80.26 & 34.77 &  6.67 & 43.87 & 25.21 & 50.54 & 39.29 & 8.50  & 71.83 & 31.70  & 39.25 & 53.01    \\
{}  & \cellcolor{gray!20} \textbf{81.72} & \cellcolor{gray!20} \textbf{41.83} & \cellcolor{gray!20} \textbf{10.33} & \cellcolor{gray!20} \textbf{48.21} & \cellcolor{gray!20} \textbf{29.04} & \cellcolor{gray!20} \textbf{53.08} & \cellcolor{gray!20}  \textbf{49.99} & \cellcolor{gray!20} \textbf{10.38} & \cellcolor{gray!20}  \textbf{73.21} & \cellcolor{gray!20} \textbf{32.42} & \cellcolor{gray!20}  \textbf{43.37}\tiny(+4.12)  & \cellcolor{gray!20} \textbf{56.31}\tiny(+3.30) \\

\arrayrulecolor{black}\hline

\multirow{2}{*}{SECOND}  & 85.84 & 50.34 &  15.18 & 58.28 & 34.41 & 66.87 & 54.73 & 25.17  &  79.94 & 57.71  & 50.59 & 62.29    \\
\arrayrulecolor{gray}\cline{2-13}
{}  & \cellcolor{gray!20}  \textbf{87.68} & \cellcolor{gray!20}\textbf{53.84} & \cellcolor{gray!20}\textbf{20.86} & \cellcolor{gray!20}\textbf{61.93} & \cellcolor{gray!20}\textbf{37.88} & \cellcolor{gray!20}\textbf{68.49} & \cellcolor{gray!20} \textbf{58.99} & \cellcolor{gray!20}\textbf{28.29} & \cellcolor{gray!20}\textbf{84.37} & \cellcolor{gray!20}\textbf{64.81} & \cellcolor{gray!20} \textbf{54.82}\tiny(+4.23)     & \cellcolor{gray!20}\textbf{65.47}\tiny(+3.18)        \\

\arrayrulecolor{black}\hline

\multirow{2}{*}{CenterPoint}      & 85.10 & 55.10 &  18.71 & 66.54 & 36.90 & 67.83 & 56.43 & 36.63  &  84.68 & 69.15   & 57.28 & 65.58     \\
\arrayrulecolor{gray}\cline{2-13}
{}  & \cellcolor{gray!20} \textbf{87.75} & \cellcolor{gray!20}\textbf{57.04} & \cellcolor{gray!20}\textbf{21.29} & \cellcolor{gray!20}\textbf{69.31} & \cellcolor{gray!20}\textbf{40.72} & \cellcolor{gray!20}\textbf{70.02} & \cellcolor{gray!20} \textbf{58.91} & \cellcolor{gray!20}\textbf{39.45} & \cellcolor{gray!20}\textbf{88.92} & \cellcolor{gray!20}\textbf{71.01}   & \cellcolor{gray!20} \textbf{60.48}\tiny(+3.20)     & \cellcolor{gray!20}\textbf{68.32}\tiny(+2.74)      \\

\arrayrulecolor{black}\hline

\multirow{2}{*}{TransFusion-L}  & 86.73 & 59.75 & 26.17 & 73.16 & 43.23 & 69.55 & 70.36 & 54.85 & 87.00 & 74.64 & 64.53 & 69.15\\
\arrayrulecolor{gray}\cline{2-13}
{} & \cellcolor{gray!20} \textbf{87.54} & \cellcolor{gray!20} \textbf{61.38} & \cellcolor{gray!20} \textbf{27.12} & \cellcolor{gray!20} \textbf{74.31} & \cellcolor{gray!20} \textbf{46.29} & \cellcolor{gray!20} \textbf{70.13} & \cellcolor{gray!20} \textbf{71.84} & \cellcolor{gray!20} \textbf{56.99} & \cellcolor{gray!20} \textbf{88.55} & \cellcolor{gray!20} \textbf{75.83} & \cellcolor{gray!20} \textbf{66.69}\tiny(+2.16) & \cellcolor{gray!20} \textbf{71.08}\tiny(+1.93) \\
\arrayrulecolor{black}\bottomrule[1pt]

\arrayrulecolor{black}\hline
\end{tabular}}}
\vspace{-4mm}
\label{tab:overall_nuscenes}
\end{table*}
where $b$ is the bounding box annotation and $b_x$, $b_y$ indicate the center coordinate of $b$. 
Inspired by~\cite{zhou2019objects},
the $\mathcal{W}(i,j,\mathbf{y})$ first calculates the probability that pixel $(i,j)$ belongs to all of the objects in ground truth $\mathbf{y}$, and then we select the largest value as the response of the pixel $(i,j)$ to the foreground object.
The $\sigma$ is a hyper-parameter determined by the object scale, which is set as $7$ in our experiments. We manually set $m=2$ and $n=3$ respectively in the decoder layers.

The overall distillation loss combines the above two parts together and can be written as:
\begin{align}
    \mathcal{L}_{sup}(\hat{\mathbf{f}}, \mathbf{f}^{*}) = \alpha  \mathcal{L}_{sup}^d(\mathbf{f}', \mathbf{f}^{*}) + \beta  \mathcal{L}_{sup}^r(\mathbf{f}^{\dag}, \mathbf{f}^{*}).
\end{align}
where we set the balance weights $\alpha=1.0$ and $\beta=0.1$.

\section{EXPERIMENTS}
\subsection{Benchmarks and Metrics}
\label{sec:Dataset}

\textbf{nuScenes}~\cite{caesar2020nuscenes} is a large-scale and popular benchmark for 3D object detection in the field of autonomous driving, which contains $700$/$150$/$150$ sequences for training, testing, and validation.
For the LiDAR-based 3D object detection task, the nuScenes dataset provides annotated point cloud data captured by LiDAR sensors. The LiDAR point clouds are represented by coordinates ($x, y, z$), reflectance value ($r$), and the time lag ($\Delta t$).
The annotations include various object categories relevant to autonomous driving, such as bicycles, buses, cars, motorcycles, pedestrians, trailers, trucks, construction vehicles, and traffic cones. For the 3D detection performance metrics, we used the mAP and nuScenes detection score (NDS) as suggested for the nuScenes dataset.

\subsection{Detectors}
\label{sec:Detector}

\textbf{PointPillars}\cite{lang2019pointpillars} is a pillar-based detector. The raw point cloud is converted to stacked pillars, which are used by the encoder to learn a set of features that can be scattered back to a 2D pseudo-image.
After the 2D backbone and FPN neck, the SSD head is employed for the prediction.
\textbf{SECOND}\cite{yan2018second} is a voxel-based detector. The raw point cloud is voxelized and fed into the SparseConv~\cite{3DSemanticSegmentationWithSubmanifoldSparseConvNet} encoder to extract 3D features, which are pooled to BEV features, and then 2D backbone and FPN module are employed, and the 2D features are used for prediction by an anchor-based head.
\textbf{CenterPoint}\cite{yin2021center} views the objects as points and introduces a center-based, anchor-free head to predict object centers, scales, and classes from BEV features.
\textbf{TransFusion-L}\cite{bai2022transfusion} uses SECOND as the backbone and designs its detection head based on a transformer decoder.
which predicts bounding boxes from a point cloud using a sparse set of object queries. 

\subsection{Implement Details}
\label{sec:Implement}
We implement our paradigm based on mmdetection3d~\cite{mmdet3d2020} with $8$ Tesla A$100$ GPUs.
For each detector, we enhance the sparse LiDAR data via the proposed Semantic Injection method and train the teacher encoder following the official configuration of the corresponding baseline~(e.g., dataset, learning rate, etc.), which inputs single-frame LiDAR data.
For the ST-Fusion method training, we employ $k=4$ frame~(2 seconds) raw LiDAR data as inputs, and the fused features are fed into the detection head for final prediction. 
Specifically, in the SA module, we set the progressively expanding kernel size as $\{1,3,5,7\}$ for the current frame and preceding ones respectively. In Semantic Supervision, the balance weights $\alpha=1.0$ and $\beta=0.1$.

\textbf{\begin{table}[]
\caption{Comparison with SOTA detectors. $^{*}$ is the reproduce method.}
\vspace{-3mm}
\centering
\begin{tabu}{ccccc}
\tabucline[0.8pt]{-}
Method & Year  & Frame & mAP & NDS \\
\hline
VoxelNeX~\cite{chen2023voxelnext} & CVPR 2023 & 1 & 60.00 &67.10 \\
LinK~\cite{lu2023link} & CVPR 2023 & 1 &63.60 & 69.50 \\
INT~\cite{xu2022int} &  ECCV 2022  & 4 &  61.80 & 67.30 \\
TM3DOD~\cite{park2024lidar} & Sensor 2024   & 4 & 63.91 & 69.94  \\
MGTANet~\cite{koh2023mgtanet} & AAAI 2023  & 4 & 64.80 & 70.60 \\
 \hline
TransFusion-L$^{*}$~\cite{bai2022transfusion} & CVPR 2022  & 1 & 64.53 & 69.15 \\
Ours & ICRA 2025  & 4 & \textbf{66.69} & \textbf{71.08} \\
\tabucline[0.8pt]{-}
\end{tabu}
\vspace{-5mm}
\label{tab:sota}
\end{table}}

\subsection{Overall Experiment Results}
\label{sec:result}
The experiment results of the detection models, including PointPillars, SECOND, CenterPoint, and TransFusion-L based on the nuScenes benchmark are illustrated in Tab.~\ref{tab:overall_nuscenes}, in which we report the overall mAP and NDS scores, along with the mAP of each class.
Our proposal significantly improves detection performance compared to the baseline models. 
For previous classical LiDAR-based detectors, such as PointPillars and SECOND, our method boosts NDS/mAP by +4.12\%/+3.30\% and +4.23\%/+3.18\% respectively, showcasing the effectiveness of our supervised fusion approach. 
Moreover, our proposal could continuously promote the detection even for state-of-the-art detectors, such as CenterPoint and TransFusion-L. Specifically, for the CenterPoint, our proposal gains +3.20\%/+2.74\% improvements over the single-frame baseline. For the TransFusion-L, our proposal could also promote the baseline model with +2.16\%/+1.93\% NDS/mAP. The huge promotion based on variant detectors demonstrates the effectiveness and universality of our proposed semantic-supervised ST-Fusion method. 
We further compare our proposal with previous state-of-the-art LiDAR-based 3D detectors as shown in Tab.~\ref{tab:sota}, including single-frame methods such as VoxelNeXt~\cite{chen2023voxelnext} and LinK~\cite{lu2023link}, as well as the sequential frame detectors INT~\cite{xu2022int}, MGTANet~\cite{koh2023mgtanet}, and TM3DOD~\cite{park2024lidar}. We could observe that with the proposed semantic supervision at the feature level, our proposal could gain superior performance against the previous state-of-the-art approaches.

\subsection{Ablation Study for Spatial-Temporal Fusion Method}
To evaluate the effectiveness of the ST-fusion module, we conduct experiments based on CenterPoint with variant multi-frame fusion strategies, including data fusion, feature fusion, and proposed ST-fusion, in comparison with the baseline that employs a single-frame input.
Specifically, data fusion concatenates multi-frame points directly, while feature fusion stacks the deep features from multiple frames extracted by the backbone.
As shown in Tab.~\ref{tab:fusion_method}, we could distinctly observe that our proposal improves the baseline model by a large margin, as well as over the data and feature fusion approaches. 
In practice, the data and feature fusion improve the baseline by +0.85\% and +1.90\% on NDS respectively, because the spatial misalignment caused by the object motion limits the potential enhancements of multi-frame data. 
For our proposal, we improve the baseline with +2.74\% NDS, as the proposed spatial-temporal fusion relieved the misalignments in spatial space, and the proposed semantic supervision fully explores its capacity.

\textbf{\begin{table}[]
\caption{Comparison with variance fusion methods.}
\vspace{-3mm}
\centering
\begin{tabu}{c|cc|c}
\tabucline[0.8pt]{-}
Method  & mAP & NDS & $\Delta$ \\
\hline
Baseline & 57.28 & 65.58 & - \\
\hline
Data Fusion & 58.86 & 66.43 & +0.85        \\
\hline
Feature Fusion & 59.79 & 67.48 & +1.90   \\
\hline
ST-Fusion~(Ours) & 	\textbf{60.48} & \textbf{68.32} & \textbf{+2.74} \\
\tabucline[0.8pt]{-}
\end{tabu}
\vspace{-3mm}
\label{tab:fusion_method}
\end{table}}

\subsection{Ablation Study for Semantic Injection Method}
In the Semantic Injection module, we enrich sparse LiDAR data by incorporating semantic labels, leading to enhanced data quality and more representative features. To evaluate its effectiveness, we compare our approach with geometric injection methods, where objects from long-term data are fused to densify the sparse LiDAR, following \cite{wang2022sparse2dense}. 
The results in Tab.\ref{tab:aug_method} show that our semantic injection method achieves superior performance compared to geometric injection~(e.g.,  +2.74\% versus +2.26\% on NDS improvements).
The reason is that the geometric injection approach suffers from incomplete object shapes and blurred surfaces caused by the fusion of moving objects, which limits the quality and representative capability of the features used for supervision.

\begin{table}[]
\caption{Comparison with variance information injection methods.}
\vspace{-3mm}
\centering
\begin{tabu}{c|cc|c}
\tabucline[0.8pt]{-}
Method  & mAP & NDS & $\Delta$ \\
\hline
Baseline & 57.28 & 65.58 & -  \\
Geometric Injection~\cite{wang2022sparse2dense} & 59.04 & 67.84 & +2.26 \\
\hline
Semantic Injection~(Ours) & \textbf{60.48} & \textbf{68.32} & \textbf{+2.74}         \\
\tabucline[0.8pt]{-}
\end{tabu}
\label{tab:aug_method}
\vspace{-3mm}
\end{table}

\subsection{Ablation Study for Module Pruning}
To assess the contribution of each module, we conduct ablation experiments by pruning the SA module, TM module, and Semantic Supervision, as detailed in Tab.~\ref{tab:module_prune}. 
The results show that both the SA and TM modules contribute to performance improvements, with each module enhancing the model's ability to process multi-frame data. Notably, the semantic supervision module delivers the most significant gains~(e.g., +1.90\% NDS), as it provides feature-level supervision that fully exploits the potential of the fusion module. 
This demonstrates the superiority of semantic guidance in improving performance by refining the fusion features.

\begin{table}[]
\caption{Module Pruning Experiment. SA and TM indicate the Spatial Aggregation and Temporal Merging modules respectively.}
\vspace{-3mm}
\centering
\begin{tabu}{c|ccc|cc|c}
\tabucline[0.8pt]{-}
Method &SA & TM & Sup & mAP & NDS & $\Delta$ \\
\hline
Baseline &  &  &  & 57.28 & 65.58 & -  \\
1 & \checkmark &  &  & 59.14 & 67.01 & +1.43 \\
2  &  & \checkmark &  & 59.29 & 67.24 & +1.66        \\
3 &  &  & \checkmark & 59.79 & 67.48 & +1.90        \\
4 & \checkmark & \checkmark &  & 59.83 & 67.28 & +1.70         \\
5 &  \checkmark &  & \checkmark & 60.01 & 67.86 & +2.28         \\
6 &  &  \checkmark & \checkmark & 60.08 & 67.94 & +2.36         \\
\hline
Ours & \checkmark & \checkmark & \checkmark & \textbf{60.48} & \textbf{68.32} & \textbf{+2.74}         \\
\tabucline[0.8pt]{-}
\end{tabu}
\vspace{-3mm}
\label{tab:module_prune}
\end{table}

\subsection{Ablation Study for Hyper-parameter Setting}
We conduct experiments for the multi-frame $k$ in E.q.~\ref{eq:tm} of the temporal merging module. 
Here, we manually set the $k$ is \{1,2,4,8\}, where the $k=1$ indicates the baseline model with single-frame input. In Tab.~\ref{tab:temporal_len}, we find that our proposal could continuously boost the detection performances while increasing the length of the input, specifically for the long-range temporal inputs. For instance, our proposal gains 68.32\% NDS with $k=4$ against 66.43\% for data fusion and 67.48\% for feature fusion.
Meanwhile, too-long sequence inputs would introduce noisy data with extreme position excursion, which could degrade the model's ability. However, with our spatial aggregate and dynamic temporal fusion, our proposal still maintains strong detection capabilities.

\begin{table}[]
\caption{Comparison with variance
temporal length $k$.}
\vspace{-3mm}
\centering
\begin{tabular}{c|cccc}
\toprule[1pt]
$K$ & 1  & 2 & 4 & 8 \\
\hline
Data-Fusion & 65.58 & 66.19 & 66.43 & 66.07 \\
Feature-Fusion & 65.58 & 66.67 & 67.48 & 66.53 \\
Ours & 66.64 & 67.62 & \textbf{68.32} & 68.29 \\
\toprule[1pt]
\end{tabular}
\label{tab:temporal_len}
\vspace{-3mm}
\end{table}

\section{CONCLUSIONS}
In this paper, we propose a novel Semantic-Supervised Spatial-Temporal Fusion method for sequential LiDAR-based 3D object detection.
Specifically, in the Spatial-Temporal Fusion module, we proposed a Spatial Aggregation method to relieve the spatial misalignment caused by the object motion over time and a Temporal Merging method to dynamically query features from the preceding frames for a comprehensive representation. 
Further, to explore the ST-Fusion capacity, we proposed a Semantic-Supervised Training~(SST) strategy, which injects the semantic information to produce high-quality features for feature-level supervision.
%
Extensive experiments demonstrate the effectiveness and superiorities of our proposed method, boosting the detector performance on the nuScenes benchmark for variant detectors. We hope our proposal will inspire future research on LiDAR-based feature fusion and representation.

\section*{ACKNOWLEDGMENT}
The work is partially supported by Shenzhen Science and Technology Program JCYJ20220818103001002, the Guangdong Key Laboratory of Big Data Analysis and Processing, Sun Yat-sen University, China, and by the High-performance Computing Public Platform (Shenzhen Campus) of Sun Yat-sen University.

\bibliographystyle{IEEEtran}
\bibliography{ref}

\end{document}